# Variable-grasping-mode gripper with different finger structures for grasping small-sized items

Tetsuyou WATANABE, *Member IEEE,* Kota MORINO, Yoshitatsu ASAMA, Seiji NISHITANI, and Ryo TOSHIMA

*Abstract*— This study presents a novel small gripper capable of grasping various types of small-sized items from flat surfaces for the assembly of small devices. Using a single actuator, the proposed gripper realizes two grasping modes: parallel-grip and turn-over modes. The gripper's mode can be switched via contact with a flat surface, such as a table. Handling thin thicknesses and light weights are the key challenges faced in attempts to grasp small-sized items. Although parallel grippers are effective in handling small items, there is a limit to the thinness of objects that can be grasped by parallel grippers. Accordingly, the turn-over mode was adopted to grasp items that exceeded this threshold. In the turn-over mode, one finger lifts the item, while another finger holds the item from above to keep it from flicking out. The proposed gripper is capable of picking up several types of items from a table, including thin (0.05 mm) and lightweight (0.007 g) items.

*Index Terms*— Gripper, small objects, grasping.

## I. Introduction

SEVERAL types of multipurpose robotic hands have been developed [1][2][3]. However, the development of robotic hands for grasping clusters of small-sized objects has not been comprehensively considered. This study addresses this challenge. Because robots are intended to support human work, the main objective of the multipurpose robot hands developed thus far is to handle items in human surroundings. The size of such items ranges from a few centimeters to 10 cm, and these items are intended to be handled by humans. Recently, small wearable devices, such as smartphones and health monitors, have become increasingly popular. To assemble such small devices, it is necessary to handle small-sized parts that are a few millimeters in size. Hence, for automation, it is desirable to develop a gripper that can handle a wide variety of items of such size.

Achieving sufficient thinness and low weight is a major challenge when developing a robot gripper that can handle items that are a few millimeters in size. A very low thickness makes it challenging to obtain a sufficiently large contact area for grasping. Although parallel grippers are effective in handling a wide variety of objects with sizes as small as a few millimeters, there is a limit to the thinness of objects that can be grasped by parallel grippers. If the object is too thin to hold, people employ their fingernails to lift them. The limitation of this scooping operation is the low weight of these objects, as they can easily fly out during scooping. Hence, there is a need for a certain type of a cover.

Considering these challenges, this study proposes a novel method for grasping small items from flat surfaces, as illustrated in Fig. 1. To handle small parts for the assembly of small products, we selected the target items presented in Fig. 2. The task here is to pick these items from a flat surface with a stable grasping posture. Both the height and width of these target items were less than 10 mm, except for a flat cable. To prove the versatility of the robot gripper, six items with different stiffnesses, dimensions, and shapes were selected. These items are typically difficult to hold using human hands. It is challenging to grasp a thin flat cable using bare hands without adopting certain tools or the edge of a desk. If the developed gripper can grasp all six items, it is expected to handle a wide variety of objects with similar dimensions.

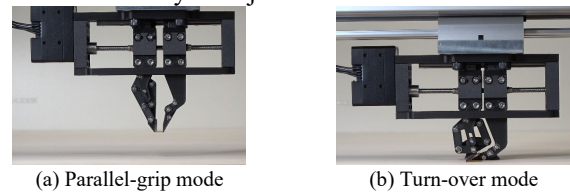

(a) Parallel-grip mode　　　　　　(b) Turn-over mode

Figure 1.　Variable-grasping-mode gripper with different finger structures

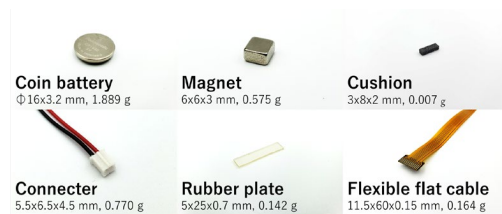

Figure 2.　Target items

The proposed gripper has two grasping modes: parallel-grip and turn-over modes. To achieve a small size and reduce costs, the two grasping modes were realized using a single actuator. The grasping mode of the gripper can be switched by

Manuscript received: February 24, 2021; Accepted: May 13, 2021; Date of current version December 12, 2019. This letter was recommended for publication by Associate Editor and Editor H. Liu upon evaluation of the reviewers' comments. This work was supported by Panasonic Corporation. (*Corresponding author: Tetsuyou Watanabe*)

T. Watanabe is with the Faculty of Frontier Engineering, Institute of Science and Engineering, Kanazawa University, Kakuma-machi, Kanazawa, 9201192 Japan (e-mail: te-watanabe@ieee.org). K. Morino, Y. Asama, S. Nishitani, and R. Toshima are with the Panasonic Corporation.

This letter has supplementary downloadable material available at http://ieeexplore.ieee.org, provided by the authors.

Digital Object Identifier (DOI): see top of this page.

contacting it with a flat surface, such as a table. We switched the grasping mode to the turn-over mode when attempting to grasp a thin item that could not be grasped in the parallel-grip mode (a precision digital scale installed on the table or a camera could be used to detect whether an item has been grasped). The turn-over mode is used for lifting while preventing the target items from flicking away. This mode is effective for thin items, and we successfully picked a 0.05-mm-thick object with a weight of 0.016 g from the table by using this mode. One finger is responsible for picking the item, whereas the other is used for holding the item from above to prevent it from flicking out. Accordingly, we developed a gripper with different finger structures on the left and right sides. A passive joint, which we termed as the scooping finger, was installed on one finger to lift items. A parallel-link-based six-bar link mechanism was installed on the other finger, such that the direction of the grasping surface could be modified via contact with a table surface as the item was held from above. We refer to this finger as the parallel-link-based passive-grasping surface variable finger. We also analyzed this six-bar link mechanism to identify the range of the force direction applied to the fingertip, which enables the switching of the grasping mode. In addition, the magnitude of the force required for switching was determined.

*A. Related works*

Examples of conventional multipurpose robotic hands are multi-fingered robotic hands that mimic the human hand, such as the DLR hand II [4][5], Barrett Hand [6][7], FRH-4 Hand [8], Pisa-IIT Hand [9][10], SoftHand Pro-D [11], ISR-SoftHand [12], UC-SoftHand [13], and Kanazawa Hand [14]. Other examples include the multi-fingered robot hand, which is specialized to primarily realize enveloping and precision grasps; Robotiq 3-Finger gripper [15]; SDM hand [16][17]; and Velo Gripper [18]. However, these hands are too large to grasp small-sized items, and as discussed earlier, the challenges pertaining to thinness and light weight cannot be addressed by simply reducing the size of these robot hands. Hence, there exists a need for special lifting motion strategies. Odhner et al. [19] developed a strategy for grasping thin objects (thicker than 1 mm); by contrast, the target range of thickness in this study was less than 1 mm. Yoshimi et al. [20] also developed a strategy for lifting thin items such as paper and plastic cards. The system proposed by Yoshimi et al. was specialized for lifting operations and could not handle other small items. Jiang et al. presented a strategy for the flex-and-flip manipulation of deformable thin objects such as paper [21]. Although these approaches are software-based, the technique proposed in this study focuses on hardware-based solutions.

Several robotic hands can lift small-sized objects, and artificial nails [20][22] are effective for lifting these objects. Nishimura et al. developed deformable fingertips filled with fluid [23], as well as an underactuated gripper with deformable fingertips and a ratchet mechanism at the joints [24]. The nails were attached to the fingertip and worked optimally to grasp thin and small objects with a scooping motion. The targets were daily objects larger than the targets adopted in this study; therefore, they did not consider the challenge posed by low weight, which causes items to fly out during grasping operations. Notably, we fabricated a smaller version of this gripper and attempted to grasp the target items shown in Fig. 2 but failed to do so. The flat cable popped out or was scooped on only one side during the grasping process. Jain et al. [25] developed a soft gripper with retractable nails that could grasp objects as thick as 0.2 mm with a success rate of 60%. However, owing to the structure of the fingernails protruding from the tips of the soft fingers, the grasping posture becomes unstable when grasping small objects. Furthermore, the target in this study was thinner and more lightweight than that used in [25].

A conveyor belt attached to the fingertip can also help realize scooping motion. Tincani et al. [26] and Ma et al. [27] developed grippers with active surfaces on the fingers to facilitate in-hand manipulation. Kakogawa et al. [28] developed a gripper with an active surface that could passively switch between grasping and pull-in operations using a single actuator with a differential mechanism. Ko developed a tendon-driven gripper with an active surface that could passively switch the actuation of the surface [29]. Morino et al. [30] developed a gripper with a surface that was passively driven by opening and closing operations via finger deflection. The concept of a gripper with a roller attached to its fingertips is similar to that of a conveyor belt gripper [31]. Although some of these demonstrated the ability to grasp thin objects such as tea bags and cloths, they primarily target daily objects and find it difficult to grasp items smaller than the curvature radius of the roller part of the fingertip, which were the items targeted in this study.

Another method for grasping thin objects is adsorption. Suction grippers are widely adopted in the industrial field and are effective in gripping thin objects; however, the surfaces of these objects must be flat and devoid of holes. In addition to this limitation, the object itself is sucked up when it is too small. The jamming gripper presented in [32] implements the same adsorption technique. The gripper consists of a soft skin (rubber bag) filled with granular materials. Hence, if the rubber bag is vacuumed, the density of the granular material is reduced, and the gripper body becomes stiff. To grasp small-sized items, a granular material smaller than the targets, as well as a very thin soft skin that can cover the targets, is required. The small jamming gripper developed by Mizushima et al. [33] succeeded in grasping an M3 bolt, which was the smallest object that could be grasped by this gripper.

The concept of adhesion is similar to that use in adsorption techniques. To grip objects, Shintake et al. used dielectric elastomer actuators to control the electroadhesion force [34]. However, the use of electrostatic forces should be avoided when handling small-sized components of small electrical products. Hawkes et al. adopted the unidirectional adhesion structure of geckos to construct a gripper with a shear grasping function [35]. Because a relatively large contact area is required for grasping with an adhesion force, it is not suitable for irregularly shaped items. The instability of the grasping posture, owing to unexpected contact, also poses a challenge.

In summary, the development of a robot hand or gripper that can grasp targeted items from flat surfaces with a stable posture has not been considered thus far. Accordingly, this study aimed to address this issue.

## II. GRIPPER FOR GRASPING SMALL-SIZED ITEMS

### A. Functional requirements

The following functional requirements were defined for this study:
1. Ability to grasp all the items illustrated in Fig. 2. from flat surfaces.
2. Presence of a mechanism to prevent items from flicking out.
3. Use of a single actuator.
4. Simple steps for the grasping technique.

The functional requirements address the challenge of grasping lightweight objects, which can fly out during grasping. We reduced the number of actuators to simplify the maintenance, management, operation, and control methods. Although it may be possible to develop a special technique for grasping a thin, lightweight object using conventional robotic hands, this study addresses the challenges involved in grasping a thin object using simple steps via a novel hardware mechanism.

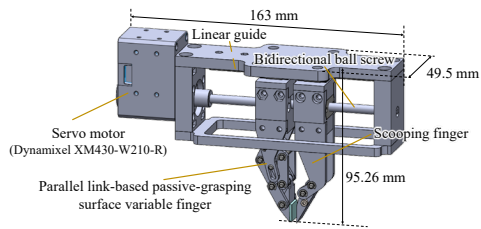

Figure 3.   Structure of the proposed gripper

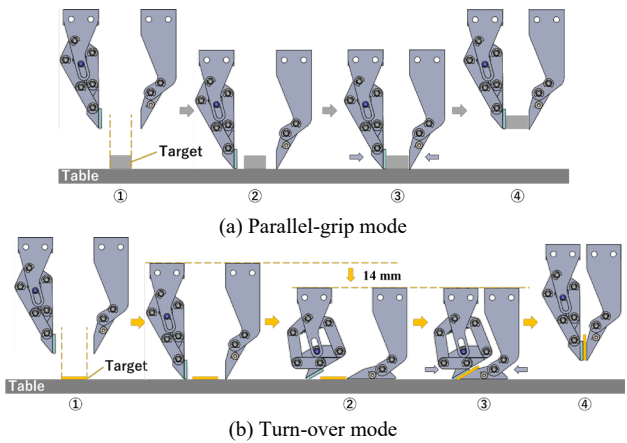

(a) Parallel-grip mode

(b) Turn-over mode

Figure 4.   Different grasping modes

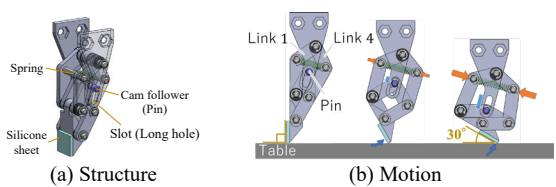

(a) Structure            (b) Motion

Figure 5.   Schematic of the parallel-link-based passive-grasping surface variable finger

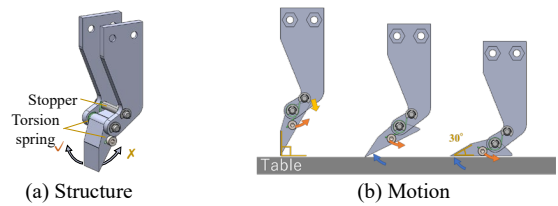

(a) Structure            (b) Motion

Figure 6.   Schematic of the scooping finger

Fig. 5 (a) presents the structure of the parallel-link-based passive-grasping surface variable finger. A six-bar linkage was installed on the finger to restrain the item from above and prevent it from being flicked out. The mechanism consists of a parallel four-link mechanism with two additional links. In the turn-over mode, the links are open, whereas in the parallel-grip mode, the links are closed. A long hole was drilled such that the movable direction of the parallel four-link mechanism was perpendicular to the direction of the hole. The links open when the fingertip comes in contact with a table, and an external force is applied to the fingertip. A spring is installed to return the opened links to their original states. As illustrated in Fig. 5 (b), the mechanism was designed such that the angle between the finger surface and the table was 30° when the links were opened to the maximum extent. In addition, the mechanism was designed such that the finger surface was perpendicular to the table when the links were closed, thus allowing it to be utilized in the parallel-grip mode. It was also designed to stop the movement of links 1 and 4 with a pin such that the link mechanism was prevented from moving when a force was applied to the fingertip in the horizontal direction, that is, when a grasping force was applied. A silicone sheet (Dragon skin 30) with a thickness of 1.5 mm was attached to the gripping surface. In the parallel-grip mode, the silicone surface took the shape of the item and contributed to achieving stable grasping. In the turn-over mode, the grasping surface rotated as the item was lifted. The silicone surface contributed toward maintaining the contact between the fingertip and the item during rotation and toward maintaining the grasp.

Fig. 6 (a) shows the structure of the scooping finger. This finger lifts an item by sliding the fingertip under the item during the turn-over operation. To accomplish this movement, we sharpened the fingertip and installed a passive joint with a torsion spring. The sharp tip of the finger rotates along the table when the finger is pressed against the table. When the finger is pressed to its maximum, the angle between the fingertip and the table is 30°, which is designed to match the surface angle of the parallel-link-based passive-grasping surface variable finger during grasping in the turn-over mode. The finger was designed such that its surface was perpendicular to the table when the finger was not pressed against the table. A stopper was installed to prevent the joint from rotating when a grasping force was applied in the parallel-grip mode.

### B. Operating procedure for grasping

The grasping technique in the parallel-grip mode is the same as that for a conventional parallel gripper (Fig. 4 (a)). The difference is that there exists an upper limit to the force that can be exerted on the table during the process. If the applied force exceeds the upper limit, the turn-over mode is activated. The upper limits are discussed in the next section. The procedure for the turn-over mode is presented in Fig. 4 (b). With the fingers open, the robot gripper is pressed against the table to the point where the parallel-link-based passive-grasping surface variable finger and the scooping finger rotate to the maximum extent (procedure 2). As the item is held from above by the parallel-link-based passive-grasping surface variable finger, the sharp tip of the scooping finger slides under the item, and the item is nipped with the tips of the two fingers (procedure 3). When the gripper is moved away from the table in this state, the fingertips and the item are rotated by 90°, such that the fingers return to their original positions, or posture, while also maintaining the grasping process (procedure 4). Once the object is lifted, the fingertip posture is the same as that in the parallel-grip mode. Although the fingertip movement differs, the grip operation procedure for grasping is almost the same as that in the parallel-grip mode, except for the amount of pressure exerted on the table.

### III. ANALYTICAL MODEL FOR PARALLEL-LINK-BASED PASSIVE-GRASPING SURFACE VARIABLE FINGER

An external force changes the grasping mode of the proposed gripper. To determine the type of external force that opens the links and switches the grasping mode, the statics involved are analyzed. The mechanism of the proposed finger can be expressed as a six-bar link mechanism, as illustrated in Fig. 7. Figs. 7 (a) and (b) present the nomenclature for the links and joints and that of the link lengths and joint angles, respectively. To derive the static relationship for the mechanism, a free body diagram is adopted. Fig. 8 shows the obtained diagrams. The origin of each free body was positioned at the joint located at the lowest position, whereas the coordinate frame was set such that the y-axis was parallel to the line connecting the joints O, T, and Q. An external force $f_e$ was assumed to be acting on the point corresponding to the fingertip. We considered $\zeta$ to be the angle between the y-axis and the direction in which $f_e$ acts. We consider $f_{ij} = (f_{ijx}, f_{ijy})^T$ ($i, j \in \{1,2, \cdots, 6\}$) to be the force applied on link $j$ by link $i$. Joints S and R are connected to the three links, and joint T is constrained by the long hole; thus, let $f_{i\iota}$ ($f_{\iota i}$) be the force applied on joint $\iota$ by link $i$ (link $i$ by joint $\iota$) ($i \in \{1,2, \cdots, 6\}, \iota \in \{S, R, T\}$) and $f_\iota$ ($\iota \in \{S, R\}$) be the driving force acting on joint $i$. Let $f_{GT}$ be the force applied on joint T, triggered by the constraint of the long hole. At points U and V, the spring force acts on links 4 and 1, respectively, and it is expressed as

$$f_k = k(l_0(\sin(\theta_0 + \theta_1) + \sin(\theta_4 + \theta_5)) - l_n) \quad (1)$$

where $k$ and $l_n$ denote the spring constant and natural length of the spring, respectively. At point O, $f_{G1}$ and $f_{G4}$ represent the forces exerted by the fixed part on links 1 and 4, respectively. When all the links are closed, as shown in Fig. 7 (corresponding to the parallel-grip mode), joints S and R are in contact with each other. In this case, links 1 and 4 are also in contact with the pin corresponding to joint T. All these contacts are lost when the link is opened. Because the purpose of this analysis is to determine the type of external force required to open the links, we investigated the force and moment balance under the condition that these contacts were lost. In addition, owing to the low weight of each link, we assumed the weight to be negligible.

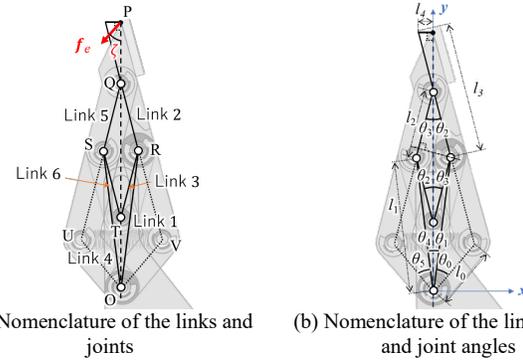

(a) Nomenclature of the links and joints   (b) Nomenclature of the link lengths and joint angles

Figure 7.  Model for the parallel-link-based passive-grasping surface variable finger

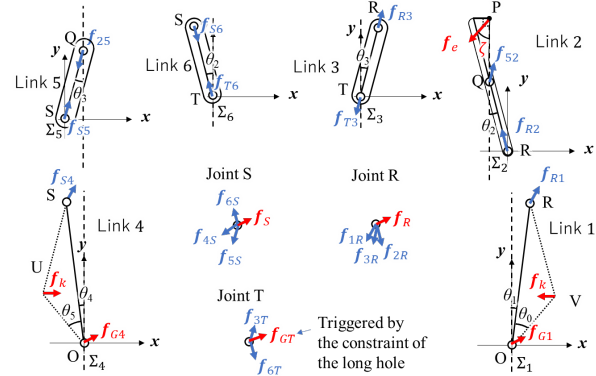

Figure 8.  Free body diagram for the parallel-link-based passive-grasping surface variable finger

In this case, the following balance equations for force and moment are obtained:

$$f_e = (-\xi \sin\zeta, -\xi \cos\zeta)^T$$
$$-f_{S5} = f_{25} = -f_{52} = (\xi\gamma \sin\theta_3, \xi\gamma \cos\theta_3)^T$$
$$f_{R2} = (\xi(\gamma \sin\theta_3 + \sin\zeta), \xi(\gamma \cos\theta_3 + \cos\zeta))^T$$
$$f_{R3} = -f_{T3} = (\beta_3 \sin\theta_3, \beta_3 \cos\theta_3)^T$$
$$f_{S6} = -f_{T6} = (\beta_6 \sin\theta_2, -\beta_6 \cos\theta_2)^T$$
$$l_1 \sin\theta_1 f_{R1y} - l_1 \cos\theta_1 f_{R1x} + l_0 \cos(\theta_0 + \theta_1)f_k \quad (2)$$
$$= 0$$
$$-l_1 \sin\theta_4 f_{S4y} - l_1 \cos\theta_4 f_{S4x} - l_0 \cos(\theta_4 + \theta_5)f_k$$
$$= 0$$
$$f_{1R} + f_{2R} + f_{3R} + f_R = 0$$
$$f_{4S} + f_{5S} + f_{6S} + f_S = 0$$
$$f_{3T} + f_{6T} + f_{GT} = 0$$

where

$$\gamma = \frac{l_4 \cos\zeta - l_3 \sin(\theta_2 + \zeta)}{l_2 \sin(\theta_2 + \theta_3)}$$

and $\xi, \beta_3$, and $\beta_6$ are the magnitudes of the forces. When the force and moment are balanced, the driving forces at joints R and S, i.e., $f_R$ and $f_S$, respectively, are zero. However, these forces were included in the relationships to evaluate the directions of $\boldsymbol{f}_R$ and $\boldsymbol{f}_S$ when the links start opening.

In this study, it is considered that joint T can only move along the y-axis owing to the constraint of the long hole. In addition, we assume that Coulomb friction acts along the y-axis, $\boldsymbol{f}_{GT} = (f_{GTx}, \mu|f_{GTx}|)^T$, where $\mu$ is the frictional coefficient. Accordingly, from (2), we obtain

$$\beta_6 = \lambda \beta_3 \tag{3}$$

where we define

$$\lambda = \frac{sign(\beta_3)\mu \sin\theta_3 + \cos\theta_3}{-sign(\beta_3)\mu \sin\theta_2 + \cos\theta_2}$$

$$sign(\beta_3) = \begin{cases} 1 (\beta_3 \geq 0) \\ -1 (\beta_3 < 0) \end{cases}$$

If (2) is aggregated using (3), we obtain

$$A\begin{pmatrix}\xi \\ \beta_3\end{pmatrix} = b - \begin{pmatrix}\cos\theta_1 & 0 \\ 0 & \cos\theta_4\end{pmatrix}\begin{pmatrix}f_{Rx} \\ f_{Sx}\end{pmatrix} \tag{4}$$

where

$$A = \begin{pmatrix}\gamma \sin(\theta_1 - \theta_3) + \sin(\theta_1 - \zeta) & \sin(\theta_1 - \theta_3) \\ \gamma \sin(\theta_3 + \theta_4) & \lambda \sin(\theta_4 - \theta_2)\end{pmatrix}$$

$$b = \begin{pmatrix}l_0/l_1 \cos(\theta_0 + \theta_1)f_k \\ -l_0/l_1 \cos(\theta_4 + \theta_5)f_k\end{pmatrix}$$

$$\xi \geq 0$$

Here, owing to the constraint of the long hole, joints S and R can only move along the $x$ direction; therefore, in (4), we only considered the forces in the $x$ direction at the joints ($f_{Rx}, f_{Sx}$). Because $\xi$ is the magnitude of the external force $\boldsymbol{f}_e$, $\xi \geq 0$. From (4), considering $f_{Rx} = f_{Sx} = 0$, $(\xi \quad \beta_3)^T = (\xi_b \quad \beta_{3b})^T$ is given under the force and moment balance as follows:

$$\begin{pmatrix}\xi_b \\ \beta_{3b}\end{pmatrix} = A^{-1}b \tag{5}$$

Here, we initially performed calculations assuming $sign(\beta_3) = 1$. When $\beta_3 < 0$, we recalculated by assuming $sign(\beta_3) = -1$.

$\xi_b < 0$ indicates that $\xi$ cannot satisfy (4); thus, the links cannot open. The force and moment balance were maintained by the constraints owing to the contact between the joints S and R, joint T and link 1, and joint T and link 4. From (4) and (5), if the magnitude of the external force is increased by $\epsilon$ (small positive constant) when $\xi_b \geq 0$ and the force and moment are balanced, the $x$ components of the driving force at the joints R and S are given by

$$\begin{pmatrix}f_{Rx} \\ f_{Sx}\end{pmatrix} = -\begin{pmatrix}1/\cos\theta_1 & 0 \\ 0 & 1/\cos\theta_4\end{pmatrix}\left(A\begin{pmatrix}\xi_b + \epsilon \\ \beta_{3b}\end{pmatrix} - b\right) \tag{6}$$

By adding $\epsilon$ under the force and moment balance, we can evaluate the directions of $f_{Rx}$ and $f_{Sx}$. The links are open only when $f_{Rx} \geq 0$ and $f_{Sx} \leq 0$. Otherwise, the contact between joints R and S and the relative configuration of joint T and links 1 and 4, that is, the contacts between them, are maintained. These contact constraints not only maintain the links but are also immobile. In summary, the links open and the grasping mode is switched to the turn-over mode when $\xi_b \geq 0$ and $f_{Rx}$ and $f_{Sx}$ satisfy (6), such that $f_{Rx} \geq 0$ and $f_{Sx} \leq 0$.

Based on the above analysis, we derived the relationship between the direction in which $\boldsymbol{f}_e$ acts on $\zeta$ and the magnitude of $\boldsymbol{f}_e$ required for opening link $\xi_b$. The parameter values adopted are presented in Table I, and Fig. 9 presents the calculated results. The area $\xi_b = 0$ indicates that the links remain closed and that the parallel-grip mode is maintained. From Fig. 9, it can be observed that the links can open when the direction of $\boldsymbol{f}_e$ is in the range of $-15°$ to $23°$. When the gripper is pressed against a table or a flat surface, the direction of the force applied by the table is $-15°(\zeta = -15°)$, which activates the transformation to the turn-over mode. If the influence of the mechanistic play is considered, $\zeta$ can be less than $-15°$. In this case, the magnitude of the force required for the transformation is approximately 5–6 N. If the direction of $\boldsymbol{f}_e$, or that of the grasping force, is in the range of $-23°$ to $90°$, the parallel-grip mode is maintained.

TABLE I. PARAMETER VALUES FOR THE ANALYSIS

| Parameter | Value |
|---|---|
| $(l_0, l_1, l_2, l_3, l_4)$ | $(10.93, 2l_2, 12.22 + l_4 \sin 15°, 2.5 \cos 15°)$ mm |
| $(\theta_0, \theta_1, \theta_2, \theta_3, \theta_4, \theta_5)$ | $(30°, 9°, 18.5°, 15°, 7.44°, 33.1°)$ |
| $l_n, \zeta, k, \mu, \epsilon$ | 9.7 mm, 15°, 0.862 N/mm, 0.6, 0.1 |

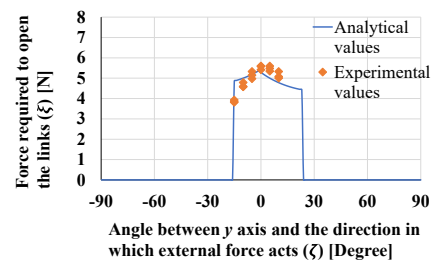

Figure 9. Relationship between the direction of external force and magnitude of the external force required to open the links

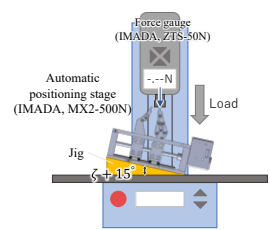

Figure 10. Experimental setup for investigating the relationship between the direction of an external force and the magnitude of the external force required to open the links

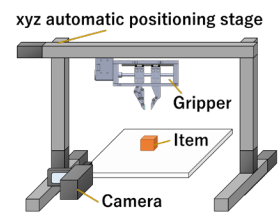

Figure 11. Experimental setup for the grasping test

To validate this analysis, we conducted an experiment in which a force gauge was used to press the fingertip in various directions. The experimental setup is illustrated in Fig. 10. To change the direction in which the force was exerted, a handmade jig with a tilted surface was attached to the base of the automatic positioning stage (Imada, MX2-500N). The fingertip was pressed by a force gauge (Imada, ZTS-50N), which was attached to the driving part of the positioning stage, at a constant speed of 10 mm/min. We recorded the force values obtained when the links started to open. We conducted three experiments for the cases $\zeta = -15°, -10°, -5°, 0°, 5°,$ and $10°$. The results are shown in Fig. 9. Except for the case of $\zeta = -15°$, the experimental values were approximately equal to the corresponding analytical values. Here, a mechanistic play exists at the joints, and the links start to move from the region of the mechanical play. By tilting the gripper, the effect of the mechanistic play was reduced. This is a primary reason for the low experimental value obtained at $\zeta = -15°$.

## IV. EXPERIMENTAL PERFORMANCE EVALUATION

### A. Grasping target items

First, we investigated whether the gripper is capable of grasping the target items shown in Fig. 2. Fig. 11 shows the experimental setup for the grasping test. In this test, the gripper was attached to the xyz automatic positioning stage (Oriental motor x-axis: ELS4XE010KD0, y-axis: ELS4YE050KD0, z-axis: EAS4LNX-E050-ARMK-2). Each target item was placed on a table and then grasped 70 times using the gripper; the corresponding grasping success rates were then derived. Note that the sample size required for the evaluation of the success rate is 68 if the margin of acceptable error is 10% and the confidence level is 90% from [36]. Success was defined as the case where the grasped item could be lifted without being dropped on the table. Fig. 12 presents the representative results obtained from the grasping test, and Table II presents the success rates. It was not possible to grasp the flat cable in the parallel-grip mode. Therefore, the turn-over mode was adopted. The flat cable was too thin to be grasped without a scooping operation. The flat cable flicked out during the scooping process when it was picked without being restrained from above. In addition, when using left and right fingers with identical structures and sharp tips (for example, using the scooping finger as both the left and right fingers), the flat cable can be scooped on one side; however, it is challenging to achieve scooping on the other side. If one side of the flat cable is lifted, the gap between the table and tip at the other side disappears; hence, it is difficult to scoop the flat cable on both sides. This phenomenon was observed in a preliminary experiment. For this reason, we adopted left and right fingers with different structures.

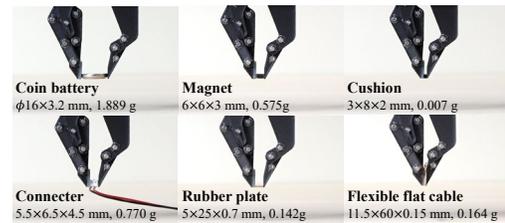

Figure 12. Representative results obtained from the grasping test of the target items

### B. Grasping other items

To evaluate the performance of the gripper, we employed it to grasp other items. This was mainly to determine the types of items that can be grasped by the developed gripper and not for a statistical analysis. Thus, the number of trials for grasping these other items was set at 10. If the success rate was neither 0% nor 100%, the number of trials was increased to 70. The turn-over mode was employed for items that could not be grasped in the parallel-grip mode, with the exception of the O-ring, plastic bag, and clip; these items were used to determine whether the gripper was able to grasp deformable and thin rigid items in the turn-over mode. When grasping cylindrical rigid items, including washers and coins, the contact area was limited. Therefore, it is difficult to grasp these items, especially if they are thin. Accordingly, we attempted to examine the minimum thickness of cylindrical rigid items that can be grasped by the developed gripper. In the parallel-grip mode, the proposed gripper was able to grasp a coin with a thickness of 1.3 mm; however, it could not grasp a spring washer with the same thickness. The surface of the spring washer was smooth, whereas the surface of the coin was jagged. This difference may have affected the results. Thus, under the parallel-grip mode, the minimum thickness of a graspable cylindrical rigid item was evaluated to be 1.3 mm (see Table III). Alternatively, the turn-over mode was used for the spring washer and M5 washer with thicknesses of less than 1.3 mm, which could not be grasped in the parallel-grip mode (the grasping of the coin is discussed in the next subsection). The success rate of grasping for the spring washer was 10/10, whereas that for the M5 washer was 61/70. This indicates that, in the turn-over mode, the minimum thickness of a graspable cylindrical rigid item is 1.0 mm. Among the items tested, the minimum thickness of a graspable item was found to be 0.7 mm (rubber plate) in the parallel-grip mode and 0.05 mm (vinyl sheet) in the turn-over mode (see Table III). These results indicate the performance limitations of the proposed gripper. The grasping of the Kim Wipe indicates that the turn-over mode can be used to grasp items with significantly low weights (0.013 g). The grasping of the PTP sheet indicates that the turn-over mode can be adopted for grasping irregularly shaped items. The grasping of the plastic bag, O-ring, and vinyl sheet indicates that the turn-over mode can be adopted for grasping deformable items. Furthermore, the grasping of the paper, heat-resistant sheet, and vinyl sheet indicates that the turn-over mode can be adopted to grasp thin items that differ from flat cables in terms of flexibility.

TABLE II. GRASPED ITEMS: NAMES OF THE TARGET ITEMS ARE SHOWN IN BOLD.

| Success rate | Parallel-grip mode | | | | | Turn-over mode | |
|---|---|---|---|---|---|---|---|
| 70/70 | **Coin battery** ($\phi$16×3.2, 1.889 g) | **Magnet** (6×6×3, 0.575 g) | **Cushion** (3×8×2, **0.007 g**) | **Connecter** (5.5×6.5×4.5, 0.770 g) | **Rubber plate** (5×25×**0.7**, 0.142 g) | **Flat cable** (11.5×60×0.15, 0.164 g) | |
| 10/10 | M6 washer ($\phi$11.5×1.5, 0.706 g) | Coin ($\phi$18.5×1.3, 2.374 g) | Resistor (2.2×6.7×2.2, 0.163 g) | Resistor (1.7×3.2×1.7, 0.126 g) | Blue belt (10×29.3×1, 0.295 g) | Spring washer ($\phi$6.6×1.3, 0.085 g) | Paper (25×76×0.1, 0.143 g) |
| | Operational amplifier (9×19.5×7, 0.960 g) | Capacitor (4.4×5×2, 0.228 g) | MOSFET (10.4×29×4.5, 1.982 g) | Electric board (18.4×25.6×4.2, 2.445 g) | Electric board (17.8×20.7×11.5, 1.869 g) | Heat-resistant sheet (10.6×40.2×0.4, 0.351 g) | Large plastic bag (66×77×4†, 3.212 g) |
| | Switch (6.3×6.3×6.8, 0.291 g) | Diode (3.5×5.5×3.5, 0.136 g) | Crimp terminal (1.95 ×2.4× 6.5, 0.054 g) | Screw ($\phi$3.3×3.3, 0.097 g) | M2 bolt ($\phi$3.8×6, 0.205 g) | Kim Wipe (13.1×49.3×0.1, **0.013 g**) | PTP sheet (16×35×4†, 0.567 g) |
| | M2.5 bolt ($\phi$4.5×6.5, 0.362 g) | M3 bolt ($\phi$5.5×7, 0.615 g) | M2 nut (4.0×4.5×1.6, 0.117 g) | M2.5 nut (5×5.5×2, 0.227 g) | M3 nut (5.4×6×2.4, 0.329 g) | Vinyl sheet (20×20×**0.05**, 0.016 g) | Plastic bag |
| | Plastic bag (25×55×4.9†, 1.787 g) | | O-ring ($\phi$8.7×1.5, 0.048 g) | | Clip (7.5×29×0.8, 0.373 g) | O-ring | Clip |
| 61/70 | | | | | | M5 washer ($\phi$10×1.0, 0.706 g) | |
| 50/70 | | | | | | Coin ($\phi$18.5×1.3, 2.374 g) | |
| 0/10 | **Flat cable** | Spring washer | Paper | PTP sheet | Large plastic bag | | |
| | Heat-resistant sheet | Vinyl sheet | Kim Wipe | | M5 washer | | |

†The thickness of the boundary area is 0.2 mm

TABLE III. PERFORMANCE OF THE PROPOSED GRIPPER

| Target property | Parallel-grip mode | Turn-over mode |
| --- | --- | --- |
| Cylindrical rigid | Thickness: ≥ 1.3 mm | Thickness: ≥ 1.0 mm |
| Overall | Thickness: ≥ 0.7 mm<br>Weight: ≥ 0.007 g | Thickness: ≥ 0.05 mm<br>Weight: ≥ 0.013 g |

## V. DISCUSSION

First, because the target items were successfully grasped with a success rate of 100%, it can be inferred that the attempted development of a gripper that meets the functional requirements was successful. Next, in Table III, we present a summary of the performance of the proposed gripper, obtained via the grasping tests. These values were evaluated using the properties of the items with the minimum thickness or weight, among the grasped items. Owing to its elasticity, the silicone surface at the fingertip conforms with the irregular shapes of small objects and contributes toward the stability of the grasp. For the scooping operation, silicone rubber was not mounted at the surface of the scooping finger; therefore, the contribution of the silicone surface toward gripping stability is limited. If an item is elastic, its surface conforms with the fingertip surface and provides a sufficiently large contact area and friction for gripping, even when silicone is not attached to the fingertip surface. This explains the gripper's ability to grasp thinner items that are elastic. When grasping cylindrical objects with thicknesses below 1.3 mm, it is challenging to obtain an adequate grasping area; hence, the grasping process in the parallel-grip mode is challenging. In such cases, the turn-over mode was found to be effective. The feasibility of grasping an item in the turn-over mode is determined by the ability of the scooping operation to push that item to the surface of the finger.

Elasticity is also effective for scooping. If an item is elastic, it can be easily lifted toward the surface of the finger. Although the risk of flicking out is high, the other finger can be used to prevent the item from flicking out by restraining it from above. Therefore, flexible and deformable items can be easily grasped in the turn-over mode. A clip and spring washers, which are examples of rigid and thin items, could be grasped in the turn-over mode. Even if an item is metallic, the lower its thickness, the more flexible it becomes. Therefore, it is possible to grasp thin and flexible materials in the turn-over mode, as demonstrated by the grasping of the flat cable. One exception was the coin, which resulted in a success rate of 100% in the parallel-grip mode but a rate of 50/70 in the turn-over mode. In one case of failed grasping, the coin penetrated the silicone surface. Compared to its front and rear, the sides of the coin result in greater friction owing to the jagged edges). This may be the reason for these unexpected results. Furthermore, it may be possible to lift the coin using a special strategy involving approaching the coin while tilting the gripper body, as reported in [20] [21]. We plan to address this issue in our future work, which aims to develop software-based strategies to overcome this limitation. The targeted connector and large plastic bag feature tail-like parts that trigger moments when grasping them. Nevertheless, a stable grasp was maintained even in the presence of these moments. As the proposed gripper is sufficiently wide to grasp small objects, this discussion primarily focused on the thinness and hardness of the target items.

The proposed gripper deals with grasping small-sized objects; hence, the maximum graspable weight is insignificant in this study. However, we observed that the fingers did not break even when an external force exceeding 30 N was applied. Because no change was observed even after repeated applications of such a large load, it can be inferred that the gripper possesses high durability. Although the accuracy of placement is not the main target, the developed gripper can place the item in its original position under both modes. Note that, in an environment where electrostatic forces are dominant, electrical grounding is required for reliable placement. The turn-over mode can be used even when the gripper is not fully pressed against a flat surface. The grasping success rate for the flat cable was 10/10 when the gripper was lifted 2, 4, 6, 8, and 10 mm from the position where it was fully pressed against the flat surface. For a lifting height of 12 mm, the success rate of grasping the flat cable was 49/70. Furthermore, the success rate was 0/10 when the lifting height was 14 mm, which corresponds to the parallel-grip mode. As the scooping angle became obtuse with an increase in the lifting height, the success rate decreased at a lifting height of 12 mm. The results also indicate that the gripper can function on surfaces with grooves or holes, even when the lowering amount of the gripper is restricted by the grooves or holes. Notably, the gripper remains functional as long as both the fingers do not get stuck inside a groove or hole during the grasping process. The turn-over mode can also be used when the surface on which the item is placed is tilted slightly. For a tilt angle of ±2.5°, the grasping success rate for the flat cable was 10/10.

## VI. Conclusion

This study presented a novel method for grasping small items, considering the task of handling small components on flat surfaces for the assembly of small products. The key challenges faced when handling such small-sized items include issues pertaining to the low thicknesses and weights of the target items. To address these issues, we developed a gripper featuring two fingers with different structures; this gripper can realize two grasping modes—parallel-grip and turn-over modes—using a single actuator. In the turn-over mode, one finger is used for scooping, whereas the other is used to hold the item from above. The modes can be switched passively via contact with a flat surface, such as a table. We analyzed the six-bar link mechanism used for the finger to identify the direction range and magnitude of the force acting on the fingertip, which is required to switch the grasping mode. The experiments demonstrated that the gripper is capable of picking several types of items from a table, including thin (0.05 mm) and lightweight (0.007 g) items. Future studies will involve the development of motion planning and assembly strategies to assemble these grasped items while constructing small devices.